\documentclass[letterpaper]{article} 
\usepackage{aaai25}  
\usepackage{times}  
\usepackage{helvet}  
\usepackage{courier}  
\usepackage[hyphens]{url}  
\usepackage{graphicx} 
\usepackage{natbib}  
\usepackage{caption} 
\usepackage{algorithm}
\usepackage{algorithmic}

\usepackage{enumerate}
\usepackage{amsthm}
\usepackage{graphicx}
\usepackage{epstopdf}
\usepackage{caption}
\usepackage{color}
\usepackage{enumitem}
\usepackage{array}
\usepackage{tabularx}
\usepackage{multirow}
\usepackage{bm}
\usepackage{booktabs}
\usepackage{tcolorbox}

\usepackage{newfloat}
\usepackage{listings}
\DeclareCaptionStyle{ruled}{labelfont=normalfont,labelsep=colon,strut=off} 
\floatstyle{ruled}
\newfloat{listing}{tb}{lst}{}
\floatname{listing}{Listing}
\title{Training With ``Paraphrasing the Original Text'' Teaches LLM to Better Retrieve in Long-context Tasks}

\author{
    Yijiong Yu\textsuperscript{\rm 1}, Zhixiao Qi\textsuperscript{\rm 1}, Zhe Zhou\textsuperscript{\rm 1}, Yongfeng Huang\textsuperscript{\rm 1}
}
\affiliations{
    \textsuperscript{\rm 1}Tsinghua University\\

    \{yuyj22, qzx21, zhouzhe20\}@mails.tsinghua.edu.cn, yfhuang@tsinghua.edu.cn
}

\UseRawInputEncoding
\begin{document}

\maketitle

\begin{abstract}
As Large Language Models (LLMs) continue to evolve, more are being designed to handle long-context inputs. Despite this advancement, most of them still face challenges in accurately handling long-context tasks, often showing the ``lost in the middle'' issue. We identify that insufficient retrieval capability is one of the important reasons for this issue. To tackle this challenge, we propose a novel approach to design training data for long-context tasks, aiming at augmenting LLMs' proficiency in extracting key information from long context. Specially, we incorporate an additional part named ``paraphrasing the original text'' when constructing the answer of training samples and then fine-tuning the model. Experimenting on LongBench and NaturalQuestions Multi-document-QA dataset with models of Llama and Qwen series, our method achieves an improvement of up to 8.48\% and 4.48\% in average scores, respectively, showing effectiveness in improving the model's performance on long-context tasks. 

\end{abstract}

\section{Introduction}
Large Language Models (LLMs) have recently emerged as top performers in a wide range of natural language processing tasks. Nevertheless, they are usually trained on segments of text of a fixed length, which means they have a predetermined context window size. Typically, their performance tends to decline markedly when the input text exceeds the context window size.

Some classic works like Yarn \cite{peng_yarn_2023}, LongChat \cite{li_dachengli1longchat_2024} and LongAlpaca \cite{chen_longlora_2023} have explored ways to make short-context LLMs better adaptable to long-context tasks. Based on these, recently, an increasing number of powerful LLMs have been equipped with the capability to handle long contexts, such as Mistral \cite{jiang_mistral_2023} and Qwen2 \cite{yang_qwen2_2024}, which can handling the text length of 32k or longer.

However, while LLMs have made remarkable progress in handling long-context tasks, as shown in many evaluations \cite{li_loogle_2023,an_l-eval_2023,wang_leave_2024}, they still lack satisfactory accuracy, and often suffer from a severe ``lost in the middle'' problem \cite{liu_lost_2023} when the context is getting longer or more complex. ``Lost in the middle'' refers to the phenomenon where the model's utilization of the context significantly weakens when the key information is located in the middle of the context. This limitation hinders the further application of LLMs in long-context scenarios.

Long-context tasks can be divided into two categories: short-dependency and long-dependency \cite{li_loogle_2023}. The former means only a small part of the long context is truly needed for the task, while the latter means most parts are needed. Because short-dependency tasks are usually more common and easier to study, for simplicity, in this paper, we start with the short-dependency tasks and mainly focus on multi-document-QA, which is one of the most representative long-context tasks and also easy to be constructed.

In short-dependency long-context tasks, the useful information in the context is sparse, thus the task can be regarded as a composite task, which can be split to ``first retrieval, then another follow-up task'' process, while the retrieval step is abstract and implicit but necessary. However, we find LLMs usually perform much worse in such composite tasks, even if they are good at each sub-task individually.

\begin{figure}[ht]
	\centering
	\includegraphics[width=1\linewidth]{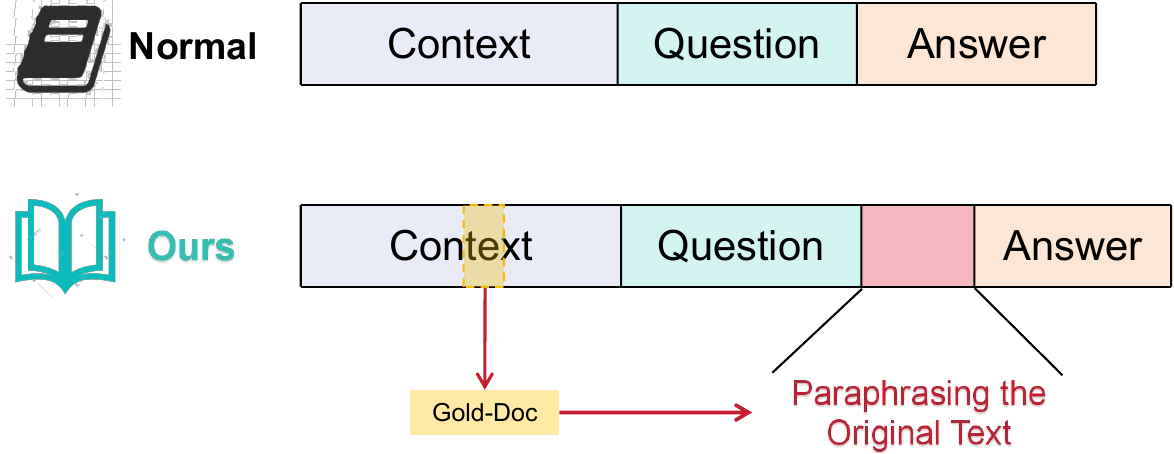}
	\caption{Our method adds ``paraphrasing the original text'' to the training samples.}
	\label{overall}
\end{figure}


In this case, it is natural to think that using CoT \cite{wei_chain--thought_2023} prompt could help. Yet, in our experiments, despite using CoT-like prompt, we find many LLMs still perform poorly, which generate wrong answers just in the first step, retrieval, even though some of them can perfectly pass ``Needle in a Haystack'' test, which represents qualified retrieval ability. While in the other hand, if they retrieve correctly, then the follow-up task becomes easier and the final answer is usually right.

We posit that this is either because these models inherently have weak retrieval capabilities, or because their retrieval abilities cannot be fully activated by just designing prompts, which hinders them from accurately locating key information in the long context. This prompts us to seek a more effective method to enhance the accuracy of models in retrieval-based composite tasks, through teaching them to better retrieve.

Thus, we propose a method based on fine-tuning with specially designed samples to enhance and activate LLMs' retrieval capability over long contexts.
Specifically, for the goal of explicitly separating and highlighting the ``retrieval'' step, when designing answers of a long-context QA sample, different from normal ways only designing a brief answer, we incorporate an additional part named ``original text paraphrasing'', which paraphrases or repeats the original text of the sentences in the context containing relevant information required for answering, as shown in Figure \ref{overall}. This part corresponds to a direct ``retrieval'' operation, while maintaining the answers' quality and coherence, which aims to not only enhance the model's retrieval capability but also teach the model to use its inherent retrieval capability more actively.

With the help of GPT-4 \cite{openai_gpt-4_2023}, we automatically construct a dataset consisting of thousands of training samples. Through evaluate models fine-tuned on the training samples designed by us, we prove our method can generally benefit the model's performance across various long-context tasks (not just retrieval or QA tasks). Besides, our method only need a light-weight dataset and a cost-effective fine-tuning stage, which is very applicable.

Our main contributions are summarized as follows:

\begin{enumerate}[itemindent = 14pt,leftmargin = 0pt]
\item We find that LLMs do not guarantee the full utilization of their retrieval capability in composite long-context tasks, even if their individual capabilities in simple tasks are strong.
\item We propose the ``original text paraphrasing'' approach, which explicitly isolates the ``retrieval'' step, to construct training samples, aiming at teaching LLMs to better use their retrieval capability.
\item Through fine-tuning and then evaluating, we prove our method can improve LLMs' retrieval capability as well as overall long-context performance.
\end{enumerate}

\begin{figure*}[!ht]
	\centering
	\includegraphics[width=\linewidth]{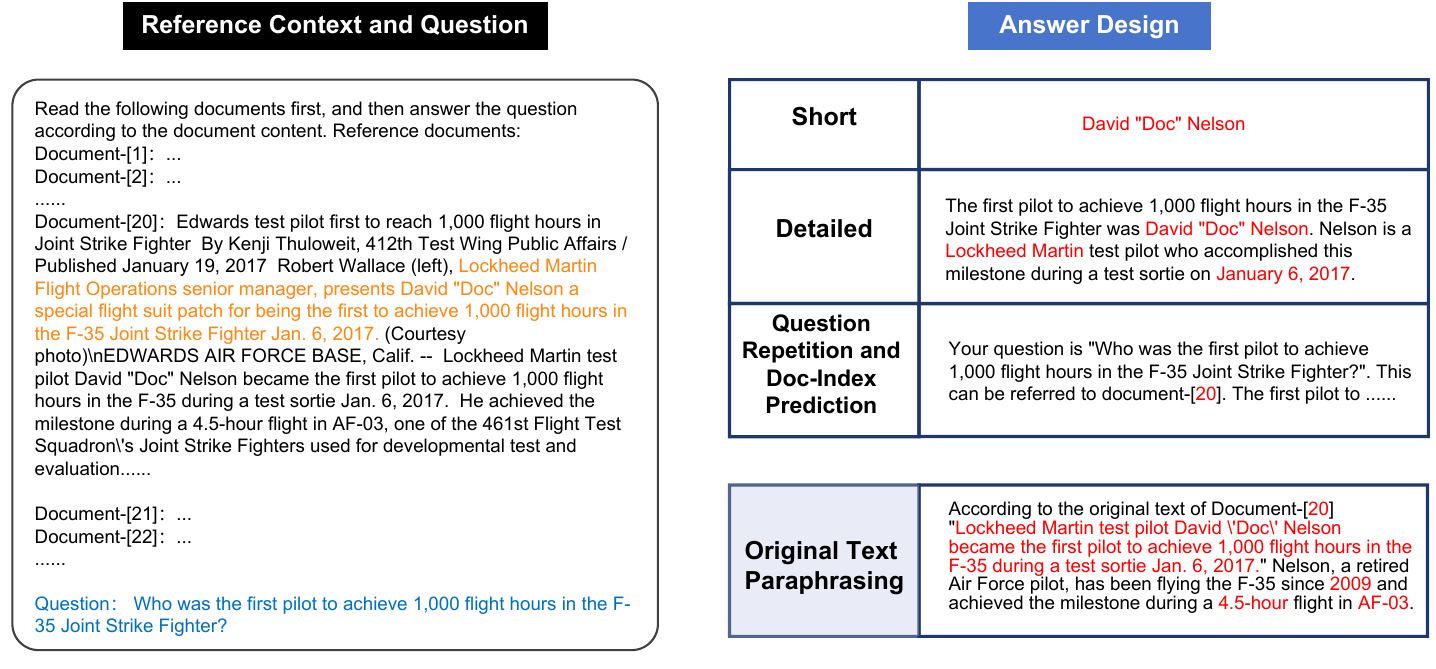}
	\caption{An example of different answer design methods for a multi-doc-QA sample. In the context and answers, key information for answering the question are highlighted.}
	\label{method}
\end{figure*}

\begin{figure*}[!ht]
	\centering
	\includegraphics[width=0.9\linewidth]{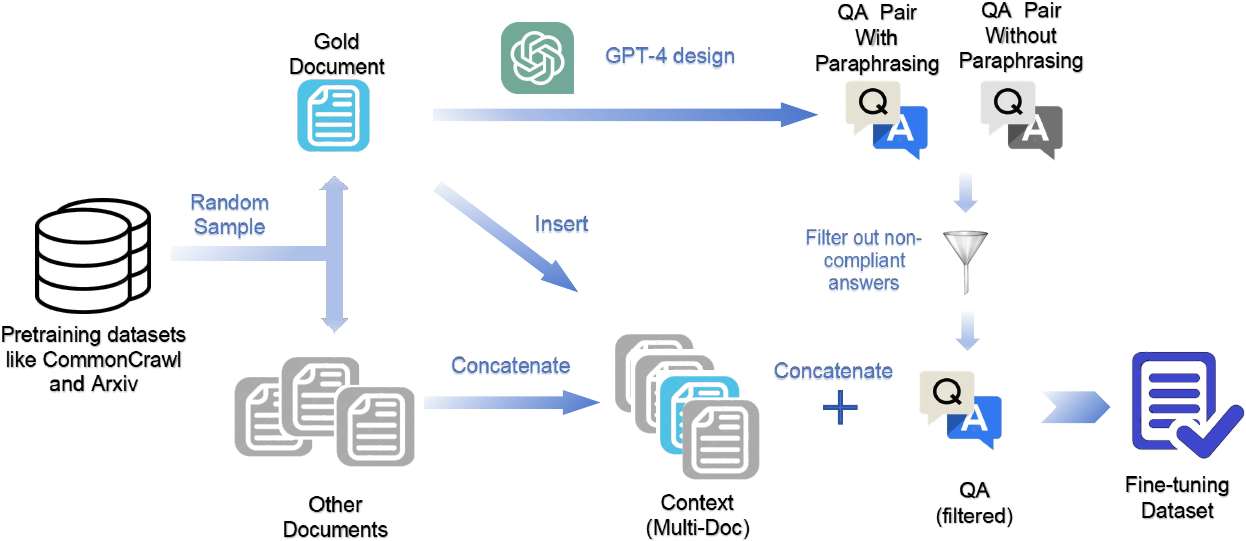}
	\caption{The pipeline of constructing our datasets with multi-doc-QA samples.}
	\label{cons dataset}
\end{figure*}

\section{Related Work}
There have been many studies that aim to improve the long-context performance of LLMs and address ``lost in the middle'' issue, which can be summarized into the following four aspects:


\subsection{Input Context and Prompt}
It is well known that using an appropriate prompt such as CoT \cite{wei_chain--thought_2023} can significantly improve the model's performance in a zero-shot way. A common method is to prompt the model to first give relevant evidence and then answer the question. This way can guide the model to first retrieve relevant information from the long text and then give answers based on this information. For example, the Claude-2.1 team proposed that adding the sentence ``Here is the most relevant sentence in the context'' \cite{anthropic_long_2023} can improve the long-context QA accuracy from 27\% to 98\%.

In addition, reorganizing the long input context is also effective. For example, LongLLMLingua \cite{jiang_longllmlingua_2023} compresses long contexts, and Attention Sorting \cite{peysakhovich_attention_2023} improves the model's utilization of contexts through reordering the input documents.

\subsection{Training Data}
Some works attempt to fine-tune models with long-context datasets to improve their long-context performance. 
Ziya-Reader \cite{he_never_2023} proposes an attention strengthening method, designing the answer with 3 parts: ``question repetition'', ``index prediction'' and ``answer summarization''. FILM \cite{an_make_2024} proposes ``Information-Intensive Training'' to teach the model that any position in a long context can contain crucial information.

\subsection{Position Embedding}
The remote attenuation characteristic of ROPE \cite{su_roformer_2022} is often considered a significant factor contributing to the ``lost in the middle'' phenomenon. Thus \citeauthor{chen_fortify_2023} propose ``Attention Buckets'', which sets up three different frequencies of ROPE and integrates them together, filling in the troughs of ROPE's remote attenuation curve. MSPOE \cite{zhang_found_2024} defines ``position-aware'' scores for each attention head and then assigns different ROPE interpolation \cite{chen_extending_2023} factors to each attention head, which significantly improves the response accuracy of Llama2 \cite{touvron_llama_2023} in long-context scenarios.

\subsection{Attention Weights}
\citeauthor{hsieh_found_2024} claim that the biased attention distribution of the model is the direct cause of its poor performance on long context, thus they decompose attention into two components, determined by position and semantics, respectively, and then calibrate the attention by eliminating the position-related component. \citeauthor{gao_empower_2023} analyze the attention matrix to directly eliminate the less important parts of attention and compensate the important parts, thereby making the attention distribution more rational.

\section{Method}
\subsection{Models Often Fail to Fully Utilize Retrieval Ability}

\begin{figure}[ht]
	\centering
	\includegraphics[width=0.9\linewidth]{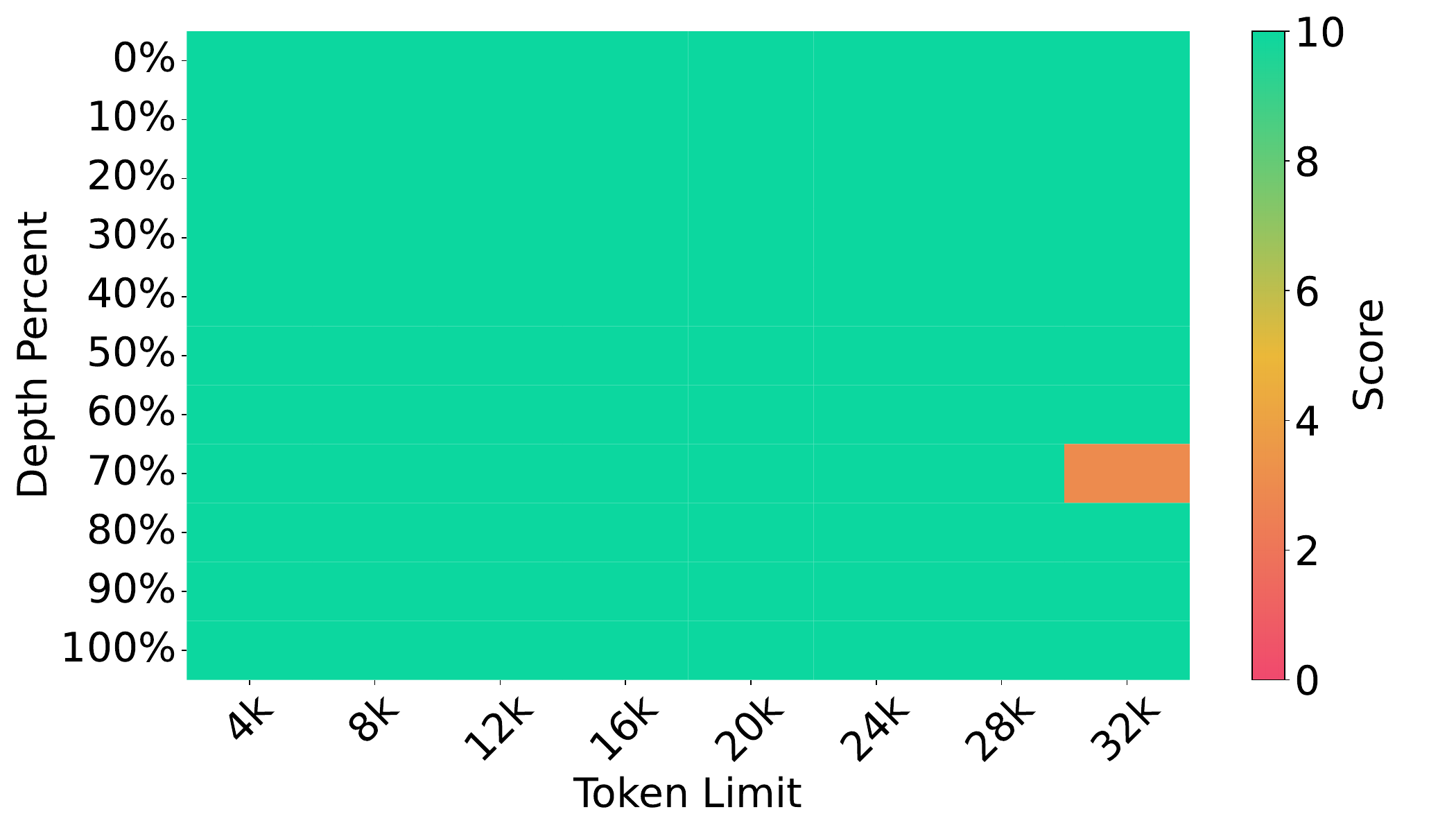}
	\caption{Qwen1.5-4b-Chat can nearly perfectly pass ``Needle in a Haystack'' test. The x-axis represents the length of the context, and the y-axis represents the position of the ``needle'' in the context.}
	\label{fig:needle qwen 4b}
\end{figure}

We take Qwen1.5-4b-Chat \cite{yang_qwen2_2024} for example, which is a model with 32k context window. As shown in Figure \ref{fig:needle qwen 4b}, it can nearly perfectly pass ``Needle in a Haystack'' \cite{gkamradt_llmtest_needleinahaystack_2023} test within 32k length, which is a task requiring LLM to retrieve a sentence containing relevant information (i.e. the ``needle'') from a long context made up of a large amount of irrelevant information, when the ``needle'' is inserted at various positions of the context. Passing this test represents a long-context LLM has the ability to directly retrieve information at anywhere of the context. 

\begin{figure}[ht]
	\centering
	\includegraphics[width=0.8\linewidth]{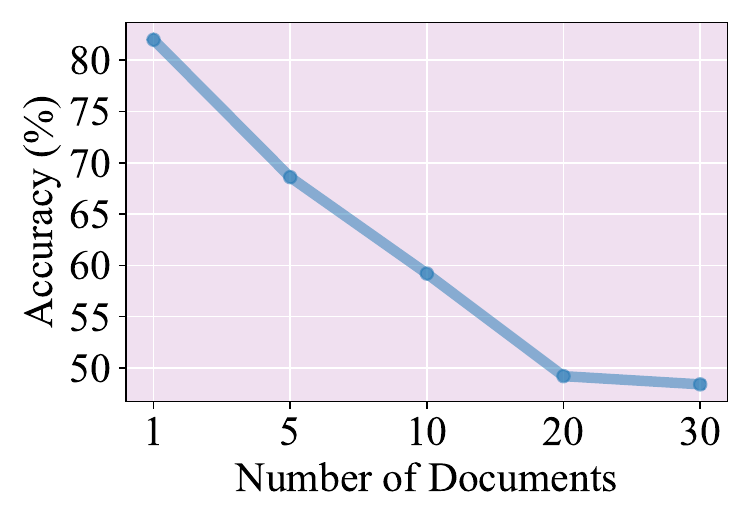}
	\caption{The accuracy of Qwen1.5-4b-Chat in multi-doc-QA task drops rapidly, as the number of documents in the context grows, with the gold document always placed in the middle position of the context.}
	\label{fig:doc num}
\end{figure}

However, as for composite tasks such as Natural Questions Multi-Document-QA \cite{liu_lost_2023} and passage retrieval tasks in Longbench \cite{bai_longbench_2023}, it performs much worse. Multi-Document-QA requires model to answer the question based on the context consisting of 20 documents with only 1 document containing truly useful information (i.e. the gold document), and passage retrieval task requires models to answer the corresponding document number given the abstract of one of 30 documents. They can both be considered as composite tasks that require both document understanding and retrieval. As shown in Figure \ref{fig:doc num}, it achieves an accuracy of 82\% in multi-doc-QA when the context only consists of the gold document, which proves its ability of understanding the document and question is fine. But when the context consists of more documents with the gold document placed in the middle, the accuracy decreases rapidly.

We have tried to awaken the retrieval capability of the model through CoT-like \cite{wei_chain--thought_2023} prompt (see Appendix for detailed prompts). However, as shown in Table \ref{tab:pass retr qwen 4b}, even though CoT-like \cite{wei_chain--thought_2023} prompt raises the accuracy in passage retrieval to some extent, the accuracy still cannot be considered satisfactory. As for Multi-Document-QA, it even cannot help.

We conclude that LLMs may not fully utilize their retrieval ability in some composite tasks, even though they perform well in simple retrieval tasks. And this issue cannot be effectively addressed by just designing more refined prompts. Therefore, to make LLMs utilize their retrieval ability more effectively in long-context tasks, we decided to fine-tuning them.

\begin{table}[htb]
    \centering
    \begin{tabular}{lcc}
    \toprule
       \textbf{Task} & \textbf{Prompt} &  \textbf{Acc}  \\
        \midrule 
        multi-doc-QA w/ 20 docs & default & 47.20  \\
       multi-doc-QA w/ 20 docs & CoT & 46.48 \\
         \midrule 
        passage retrieval & default & 13.00  \\
       passage retrieval & CoT & 26.00\\
    \bottomrule
    \end{tabular}
    \caption{Performance of Qwen1.5-4b-Chat on Natural Questions Multi-Document-QA and passage\_retrieval\_en of LongBench with different types of prompts.}
  	\label{tab:pass retr qwen 4b}
\end{table}

\subsection{Original Text Paraphrasing}

We have identified that an accurate ``retrieval'' step is crucial for completing the whole task. To explicitly isolate and emphasize the ``retrieva'' process, while also ensuring that the retrieved content aids in the better completion of subsequent tasks, when constructing long-context training samples' answers, we design a ``paraphrasing the original text'' task at the beginning part of the answer, and the remaining part of the answer is crafted based on the question and the paraphrased original text to provide a final response. In ``paraphrasing the original text'' task, we extract and then repeat the original text of the most crucial parts (usually several sentences directly related to answering the question) of the context.

Figure \ref{method} illustrates how we design an multi-doc-QA sample with ``original text paraphrasing'', which is compared with other methods: ``short'' and ``detailed'' are traditional ways used by Multi-passage-QA-from-Natural-Questions \cite{together_togethercomputerlong-data-collections_2023}, ``question repeating and doc-index prediction'' is proposed by Ziya-Reader \cite{he_never_2023}. The most crucial part in the reference documents for answering the question has been highlighted, and correspondingly, the words or sentences in the answer that pertain to this part have also been highlighted. It can be clearly seen that our design is capable of encompassing the maximum amount of key information, not only emphasizing the retrieval process to the greatest extent but also enabling the subsequent generation to directly and effectively utilize the retrieved information.

\subsection{Dataset Construction}
We leverage the power of GPT-4 (gpt-4-0613) \cite{openai_gpt-4_2023} to generate high-quality in-context question-answer pairs according to our requirements. Figure \ref{cons dataset} shows how we build multi-doc-QA samples with ``original text paraphrasing''. For English dataset, we randomly select about 100k samples from the CommonCrawl dataset \citep{patel_introduction_2020} as reference documents for constructing the context, and for Chinese, Wudao 200GB open-source Chinese corpus \citep{baai_wudao_2023} is used. Then, we let every 100 reference documents form a group, and take the first document from each group as the gold document, based on which, GPT-4 \citep{openai_gpt-4_2023} designs a question-answer pair under the prompt that the relevant original text must be provided in the answer, as follows:

\begin{tcolorbox}[colback=white]
Document:\\
\{document content\}\\
Please degisn an question answer pair base on the document. \\
The answer of the QA pair must start with ``According to the original text `......' '', first give the relevant original text in the reference content, and then answer the question in detail.
\end{tcolorbox}

If the designed answer does not start with ``According to the original text `......' '', we let GPT-4 generate again until it meets our requirements.

The remaining documents are distractors. To control the context length not exceeding 32k, we randomly discard a certain number of distractor documents. Then, all the documents in the group are randomly shuffled and concatenated to form a single context containing these documents, with each document preceded by its corresponding serial number. The question designed by GPT-4 is appended at the end of the context, with the answer designed by GPT-4 serving as the ground-truth answer. In addition, to make the answer more complete, we add the relevant document's number before the ``original text paraphrasing'' part, like ``According to the original text of document [5] `......' '', so that the answer contains both ``document index predicting'' and ``original text paraphrasing'' tasks.

For constructing single-doc-QA samples, we use papers from Arxiv and CNKI as the context, and for each paper, we randomly select one paragraph based on which GPT-4 designs a QA-pair. Other operations are the sample as multi-doc-QA.

In fact, we originally used gpt-3.5-turbo, but when transferred to gpt-4, we found the overall performance was not improved much. Therefore, although intuitively using a more advanced LLM should be better, our method does not necessitate a very strong LLM or a specific version, as long as it can follows the format requirement in most cases.

Following LongAlpaca dataset\cite{chen_longlora_2023}, to maintaining models' generalization, we also add 2,000 short-form instruction samples, which are sampled from Alpaca \cite{rohan_taori_stanford_2023} dataset and Llama2-chinese dataset \cite{flagalpha_flagalphallama2-chinese_2023}, into our dataset. Finally, our dataset comprises 2,000 short-form samples and 3,000 long-context samples with ``original text paraphrasing'' that we synthesized, with an equal number of Chinese and English samples. The length of the long-context samples varies from 8k to 32k and 80\% of them are multi-document-QA samples while the rest is single-document-QA.

\section{Experiments}
\label{sec:exp}
\subsection{Implementation Details}
We use models in Qwen series \cite{bai_qwen_2023} and the Chinese version of Llama series \cite{touvron_llama_2023}, which support both English and Chinese, as our base models for fine-tuning on our dataset.

Using the Qlora \cite{dettmers_qlora_2023} training method, we quantized the base model to int4 and fine-tuned all the linear layers of the models, with the following parameters: lora rank is 16, lora alpha is 64, global batch size is 16 and learning rate starts from 1e-5 with linearly decreasing. During fine-tuning, only the answer part was set as labels and involved in the loss calculation. Each model was trained for 1 epoch on a single A100 GPU, taking about 10 hours. 

As an ablation experiment, we removed the ``original text paraphrasing'' part in the answers of our dataset while keeping other parts unchanged, and again fine-tuned the model as a contrast (SFT w/o ours). We also construct a dataset with the same size using Ziya-Reader's methods \cite{he_never_2023} as a comparable baseline.

In evaluation, we use greedy decoding strategy for all the models and all the datasets to ensure reproducibility, and the model precision is set to bfloat16.

\begin{table*}[ht]
    \centering
    \begin{tabular}{l|cccccc}
    \toprule
        Models & MultiDoc & SingleDoc & Synthetic & Summurization & FewShot & AVG \\
         \midrule 
Qwen1.5-4b-Chat & 65.06 & \textbf{77.70} & 14.75 & 19.27 & \textbf{57.00} & 46.76 \\
Qwen1.5-4b-Chat SFT w/ ours & \textbf{65.39} & 72.22 & \textbf{62.00} & 19.57 & \textbf{57.00} & \textbf{55.24} \\
Qwen1.5-4b-Chat SFT w/o ours & 65.28 & 71.86 & 12.75 & \textbf{19.84} & 56.50 & 45.25 \\
Qwen1.5-4b-Chat SFT w/ Ziya & 65.32  & 76.28  & 25.75 & 19.50 & 56.88 & 48.75  \\
         \midrule 
Qwen2-7b-Instruct & 60.89 & 80.32 & 63.00 & 19.62 & 60.00 & 56.77 \\
Qwen2-7b-Instruct SFT w/ ours & 65.80 & 82.17 & \textbf{65.00} & \textbf{20.34} & 61.50 & \textbf{58.96} \\
Qwen2-7b-Instruct SFT w/o ours & \textbf{66.49} & 82.12 & 56.75 & 16.98 & 61.75 & 56.82 \\
Qwen2-7b-Instruct SFT w/ Ziya & 65.42 & \textbf{82.40}  & 61.50 & 19.88 & \textbf{62.50} & 58.34 \\
         \midrule 
Llama3-8b-chinese-Chat & 64.04 & 79.20 & 80.75 & 18.56 & 50.62 & 58.63 \\
Llama3-8b-chinese-Chat SFT w/ ours & 65.22 & \textbf{80.92} & \textbf{97.75} & \textbf{21.17} & \textbf{57.38} & \textbf{64.39} \\
Llama3-8b-chinese-Chat SFT w/o ours & \textbf{66.21} & 80.62 & 97.00 & 20.64 & 57.12 & 64.32 \\
Llama3-8b-chinese-Chat SFT w/ Ziya & 63.80  & 80.90 & 68.75 & 19.44 & 51.75 & 56.93 \\
    \bottomrule
    \end{tabular}
    \caption{Performance of different training methods with different models on LongBench~\cite{bai_longbench_2023}.}

    \label{tab:long_bench}
\end{table*}

\begin{table*}[ht]
    \centering
    \begin{tabular}{l|cccccc}
    \toprule
        Models & 1st & 5th & 10th & 15th & 20th & AVG \\
         \midrule 
Qwen1.5-4b-Chat          & \textbf{57.60} & 46.60 & 45.20 & 43.80 & 42.80 & 47.20 \\
Qwen1.5-4b-Chat SFT w/ ours & 55.40 &  \textbf{49.40} &  \textbf{60.00}   & \textbf{54.80} & 38.80 & \textbf{51.68} \\
Qwen1.5-4b-Chat SFT w/o ours & 57.00 & 47.00 & 48.00 & 43.60 & 44.80 & 48.08 \\
Qwen1.5-4b-Chat SFT w/ Ziya & 52.20 & 46.60 &  44.60 & 43.00  &  \textbf{51.00}  &  47.47 \\
         \midrule 
Qwen2-7b-Instruct           & 72.20 & 56.00 & \textbf{55.00} & 55.40 & \textbf{57.40} & 59.20 \\
Qwen2-7b-Instruct SFT w/ ours & \textbf{78.40} & \textbf{59.40} & \textbf{55.00} & 52.20 & 56.60 & \textbf{60.32} \\
Qwen2-7b-Instruct SFT w/o ours & 70.20 & 57.80 & 53.40 & 53.40 & 51.40 & 57.24 \\
Qwen2-7b-Instruct SFT w/ Ziya & 75.40 & 58.80 &  54.20 &  \textbf{56.00} &  54.60 &  59.80  \\

         \midrule 
Llama3-8b-chinese-Chat             & 69.00 & 61.80 & 61.20 & 58.40 & 64.00 & 62.88 \\
Llama3-8b-chinese-Chat SFT w/ ours & \textbf{71.20} & \textbf{63.60} & \textbf{64.60} & 60.80 & 64.60 & \textbf{64.96} \\
Llama3-8b-chinese-Chat SFT w/o ours & 69.00 & 61.40 & 63.40 & \textbf{62.00} & \textbf{67.80} & 64.72 \\
Llama3-8b-chinese-Chat SFT w/ Ziya & 69.80 & 60.60 & 61.40 & 58.80 & 66.40 & 63.40  \\

    \bottomrule
    \end{tabular}
    \caption{Performance of different methods with different models on NaturalQuestions Multi-Doc-QA \cite{liu_lost_2023} when the gold document is placed at 5 different positions (at 1st, 5th, 10th, 15th, 20th with 20 documents in total).}
    \label{tab: lost middle}
\end{table*}

\begin{table*}[htb]
    \centering
    \begin{tabular}{lc|cccccc}
    \toprule
       Prompt & Method & 1st & 5th & 10th & 15th & 20th & Avg  \\
         \midrule 
\multirow{2}{*}{CoT} & baseline & \textbf{59.80} & 42.80 & 44.60 & 42.60 & \textbf{42.60} & 46.48 \\
                   &  SFT w/ ours & 54.20 & \textbf{51.00} & \textbf{59.40} & \textbf{51.80} & 36.20 & \textbf{50.52} \\
         \midrule 
\multirow{2}{*}{not CoT} & baseline & \textbf{56.80} & 42.40 & 42.80 & 41.00 & \textbf{41.00} & 44.79 \\
& SFT w/ ours & 52.60 & \textbf{46.60} & \textbf{52.60} & \textbf{44.00} & 36.20 & \textbf{46.39} \\

    \bottomrule
    \end{tabular}
    \caption{Performance of Qwen1.5-4b-Chat on NaturalQuestions Multi-Doc-QA \cite{liu_lost_2023}, when using different prompts. The gold document is placed at 5 different positions respectively (at 1st, 5th, 10th, 15th, 20th with 20 documents in total).}
  	\label{prompt type}
\end{table*}

\begin{table}[htbp]
\centering
\begin{tabular}{l|cc}
    \toprule
Model & 0 shot & 5 shot \\
         \midrule 
Qwen1.5-4B-Chat & 25.77 & 52.83 \\
Qwen1.5-4B-Chat SFT w/ paraph & 52.57 & 51.16 \\
         \midrule 
Qwen2-7B-Instruct & 70.82 & 70.96 \\
Qwen2-7B-Instruct SFT w/ paraph & 70.66 & 70.75 \\
         \midrule 
Llama3-8B-Chinese-Chat & 64.94 & 67.22 \\
Llama3-8B-Chinese-Chat SFT w/ paraph & 63.83 & 65.85 \\
    \bottomrule

\end{tabular}
\caption{Model performance in MMLU with 0-shot and 5-shot settings.}
\label{tab:mmlu}
\end{table}

\begin{figure}[ht]
	\centering
	\includegraphics[width=0.9\linewidth]{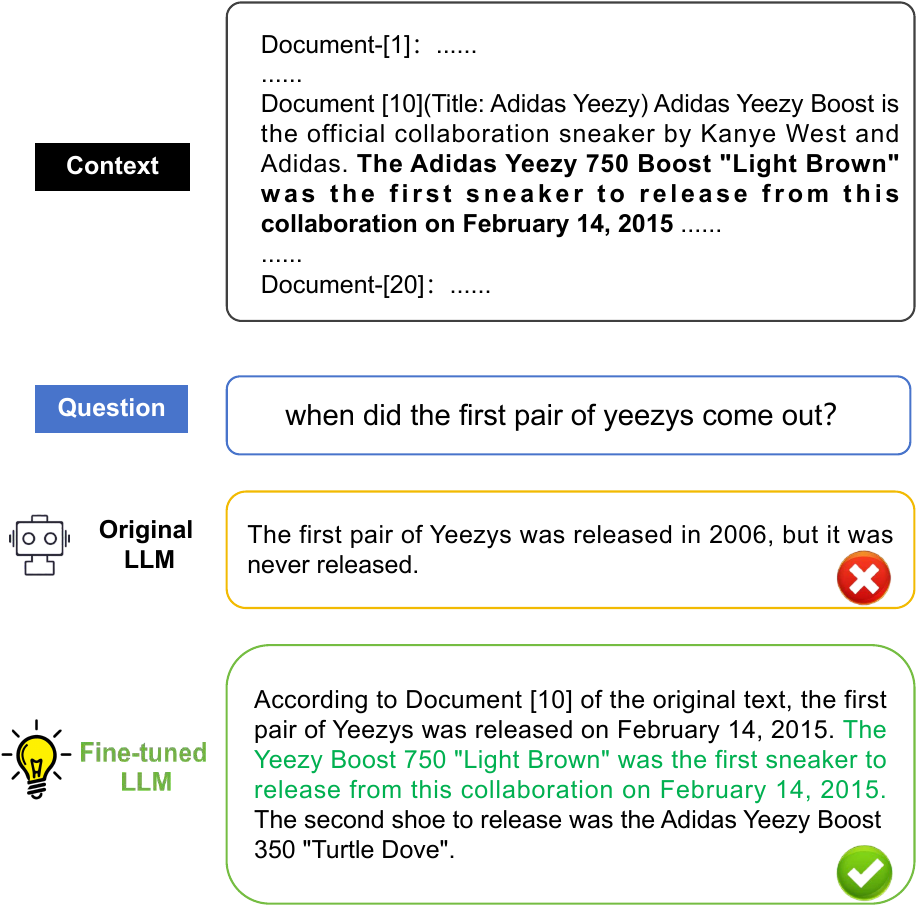}
	\caption{A case showing how the fine-tuned model (Qwen1.5-4b-Chat) gives a better answer in multi-doc-QA task. The key sentence is highlighted.}
	\label{fig:right case}
\end{figure}

\subsection{Evaluation}
\subsubsection{Evaluation on Long-context Tasks}

We evaluated our model on LongBench \cite{bai_longbench_2023}, a commonly used bilingual and multi-task dataset for long-context evaluation. It contains 14 English tasks and 5 Chinese tasks, across 5 task types: multi-doc-QA, single-doc-QA, synthetic tasks, summarization and few-shot learning. For each task type, to ensure a balanced sample size for both languages, we choose a English sub-dataset and a Chinese sub-dataset (specific dataset selection is in Appendix). Every sub-dataset contains 200 samples. 



We evaluate our models on these 5 types of long-context tasks. For synthetic (i.e. passage retrieval) and few-shot tasks, we use exact-match to calculate accuracy and for summarization tasks, we use rouge-L. For multi-doc-qa and single-doc-qa tasks, we also use accuracy (see Appendix for how to calculate accuracy for QA). To avoid OOM, the maximum input length is set to 16k for all models.

The evaluation results are shown in Table \ref{tab:long_bench}. Our method achieves the best average performance, and is particularly outstanding on the retrieval tasks (Synthetic), even if our dataset does not include an specialized retrieval task but just dissociates the implicit retrieval process in in-context QA tasks into an explicit process, which proves that adding ``original text paraphrasing'' to the training samples does significantly improve the model's retrieval capability. 

However, the improvement on single-document-QA tasks is unstable, which may be due to the relatively short sample length of such tasks in Longbench (about 5k in average length), making it difficult to consistently reflect the model's long-context capability. 

\subsubsection{Evaluation on ``Lost in the middle'' Issue}

The ``lost in the middle'' phenomenon \cite{liu_lost_2023} represents a prevalent issue for LLMs when processing long-context tasks, signifying a low efficiency in utilizing information in the middle of the context. We use the multi-doc-QA datasets provided by ``lost in the middle'' research \cite{liu_lost_2023}, which contains 2,655 samples. For each test sample, its context comprises a total of 20 documents, with only one document containing the information corresponding to the question attached to the context. The key document is placed at positions 1st, 5th, 10th, 15th, and 20th, respectively, to evaluate the model's performance when the key information is located at different positions within the context. We also use accuracy (see Appendix) as the metric and the prompt is the same as the original paper.

The results in Table~\ref{tab: lost middle} shows, our method achieves the highest average score on this dataset, and in most cases, it demonstrates great improvement when the key document is located in the middle. This substantiates that our method to fortify the model's retrieval capabilities provides an effective way to mitigate the 'lost in the middle' issue and make LLMs more effectively utilize the information in the middle of the long context.

\subsubsection{Case Study}
An example of how our fine-tuned model generates a more accurate answer in multi-doc-QA is shown in Figure \ref{fig:right case}. The original model seems to have completely failed to retrieve the correct paragraph, and gives a wrong answer, even though it has the context window of 32k length, which is much larger than the prompt length, 4k. In contrast, the fine-tuned model can not only correctly give the document number, but also retell the relevant original text as a reference (although there is a slight discrepancy in the word order and format of the model's responses compared to our training samples), which proves it has learned to better utilize its retrieval skills to give a more accurate answer.

\subsubsection{Influence of Prompt Style}
By incorporating ``original text paraphrasing'' into our training data, it may change the format of the model's responses to a CoT-like \cite{wei_chain--thought_2023} form after fine-tuning. We want to find out, if during inference, no special response format is used and we ask the model to provide answers straightforwardly, will our method still show improvement? Thus we use 2 opposite prompts to control model's response format: 1. CoT-like: we require the model to provide reference text before answering; 2. not CoT-like: we ask the model to provide the answer straightforwardly and concisely.

As shown in Table \ref{prompt type}, the model trained with our method achieves higher average scores in both prompts. Moreover, though CoT-like prompt can generally improve accuracy, with our fine-tuning, the model can benefit more from CoT-like prompt (the improvement of CoT is 4.13\% for the fine-tuned model and 1.69\% for the base model). This demonstrates that our method does not merely alter the format of the model's responses, but rather enhances the intrinsic capabilities of the model.

\subsubsection{Evaluation on General Ability}
Most training samples of our datasets are long-context, so we want to test if such training will impair models' performance in general tasks, which are usually short-context (degrading in short-context tasks is a common trade-off for long-context training) . We evaluate the fine-tuned models on MMLU benchmark \cite{hendrycks_measuring_2021}, as shown in Table \ref{tab:mmlu}. Although there is a slight decrease in the score in most cases, the maximum decline does not exceed 2\%, which remains within an acceptable range. Hence it can be regarded that our method substantially does not impair the model's generalization ability.

\section{Conlusion}
In this paper, we propose a novel and practical method to enhance model's long-context capability. We first discover models often fail to correctly retrieval in composite tasks. Then, we use ``paraphrasing the original text'' to construct a training dataset which can effectively enhancing model's retrieval capability. Finally, through low-cost fine-tuning, we obtain models with superior performance on long-context tasks. Our research provides a new way to help LLMs handle long context better, and mitigate the problem of ``lost in the middle''. 

\section{Acknowledgments}
This work was supported by National Natural Science Foundation of China (U2336208,82090053,61862002).

We are grateful to Huiqiang Jiang from Microsoft Research Asia for his guidance in writing and revising this paper.


\bibliography{mybib}

\appendix
\section{Appendix}
\label{appendix}

\subsection{Datasets in Longbench}
\label{app:longbench dataset}
The datasets we selected is shown in \ref{tab:datasets longbench}. For each task type, we choose one English dataset and one Chinese dataset, and the score of this task is averaged across these two.

\begin{table}[h]
\centering
\begin{tabular}{l|ll}
\toprule
\textbf{Task} & \textbf{dataset en} & \textbf{dataset zh} \\ \midrule
Multi-doc-QA & hotpotqa & dureader \\
Single-doc-QA & multifieldqa & multifieldqa \\ 
Retrieval & passage\_retrieval & passage\_retrieval \\ 
Summarization & qmsum & vcsum \\ 
Few-shot & trec & lsht \\ \bottomrule
\end{tabular}
\caption{Datasets names for different tasks in Longbench.}
\label{tab:datasets longbench}
\end{table}

\subsection{Metric for QA Tasks}
\label{best subspan}
We use the same metric as which has been used in ``lost in the middle'' issue \citep{liu_lost_2023}. This metric evaluate whether any of the correct answers appear in the predicted output. The calculation is very simple. If the ground truth (usually a few words) is in the model's prediction, it is considered a correct sample, otherwise it is incorrect, and the accuracy is calculated accordingly. Before calculating, punctuation and articles will be removed from the text, and all letters will be lowercase.

\subsection{The CoT-like prompt}
\label{app:cot}
The default prompt for multi-doc-QA task is:

\begin{tcolorbox}[colback=white]
Write a high-quality answer for the given question using only the provided search results (some of which might be irrelevant). \\
\{documents\}\\
Question: \{question\} \\
Answer:
\end{tcolorbox}

We use 2 opposite prompt in multi-doc-QA task to test if the improvements of our methods are sensitive to prompt method such as CoT \cite{wei_chain--thought_2023}. The first is CoT-like:

\begin{tcolorbox}[colback=white]
Write a high-quality answer for the given question using only the provided search results (some of which might be irrelevant). \\
When answering, \textbf{you must first paraphrase the original text of the relevant paragraphs or sentences of the provided search results.}\\
\{documents\}\\
Question: \{question\} \\
Answer:
\end{tcolorbox}

And the other prompts the model not to use CoT:

\begin{tcolorbox}[colback=white]
Write a high-quality answer for the given question using only the provided search results (some of which might be irrelevant). \\
When answering, \textbf{you mustn't repeat paragraphs or sentences in the provided results, but give the answer directly and concisely.}\\
\{documents\}\\
Question: \{question\} \\
Answer:
\end{tcolorbox}

In passage retrieval task of Longbench \cite{bai_longbench_2023}, the default prompt is 

\begin{tcolorbox}[colback=white]
Here are 30 paragraphs from Wikipedia, along with an abstract. Please determine which paragraph the abstract is from.\\
\{context\}\\
The following is an abstract.\\
\{input\}\\
Please enter the number of the paragraph that the abstract is from. The answer format must be like ``Paragraph 1'', ``Paragraph 2'', etc. The answer is:
\end{tcolorbox}

And we also experiment with a CoT-like prompt:

\begin{tcolorbox}[colback=white]
Here are 30 paragraphs from Wikipedia, along with an abstract. Please determine which paragraph the abstract is from.\\
\{context\}\\
The following is an abstract.\\
\{input\}\\
\textbf{Let's think step by step. You must first analyse the abstract and the relevant paragraph,} and then give the number of the paragraph that the abstract is from.
\end{tcolorbox}

\subsection{Time Consumption}
Although our method needs to generate more text, in practice, our method does not increase the time cost too much in long-context scenarios. As shown in Table \ref{tab:time}, we compare the inference time cost of our method and the baseline, which indicates there is only a minor increase on time consumption when the context is long.

\begin{table}[h]
\centering
\begin{tabular}{l|ll}
\toprule
\textbf{prompt length} & \textbf{baseline} & \textbf{ours} \\ \midrule
3k &	1.4 &	2.5  \\
24k &	12.4 &	15.1  \\

\bottomrule
\end{tabular}
\caption{The average time consumption (seconds) per QA sample, in different input lengths, in the Multi-Doc-QA task.}
\label{tab:time}
\end{table}



\subsection{Why Previous Training Sample Design is not Optimal: A Conjecture}
\label{ziya}
Ziya-Reader \citep{he_never_2023} proposed question repetition and index prediction tasks. For the question repetition task, since the fixed position of the question is either at the beginning or end of the context, it is not a segment that needs to be retrieved, hence the task has no contribution to retrieval relevance. In the index prediction task, although the index of the document is directly related to key segments within the context, and thus highly retrieval-relevant, its contribution to the loss in training samples is minimal due to its very short length, which hardly affects overall improvements. 
\pagestyle{empty}

\end{document}